# Non-Minimal Triangulations for Mixed Stochastic/Deterministic Graphical Models

**Chris D. Bartels and Jeff A. Bilmes**
{bartels,bilmes}@ee.washington.edu
University of Washington, Department of Electrical Engineering
Seattle, WA 98195

## Abstract

We observe that certain large-clique graph triangulations can be useful for reducing computational requirements when making queries on mixed stochastic/deterministic graphical models. We demonstrate that many of these large-clique triangulations are non-minimal and are thus unattainable via the elimination algorithm. We introduce *ancestral pairs* as the basis for novel triangulation heuristics and prove that no more than the addition of edges between ancestral pairs need be considered when searching for state space optimal triangulations in such graphs. Empirical results on random and real world graphs are given. We also present an algorithm and correctness proof for determining if a triangulation can be obtained via elimination, and we show that the decision problem associated with finding optimal state space triangulations in this mixed setting is NP-complete.

## 1 INTRODUCTION

When expressing a probability distribution as a Bayesian network, it is often beneficial to include variables whose value is a deterministic function of other variables. These variables have a variety of possible uses, such as the representation of hard or soft constraints [10]. One might also use a hidden deterministic variable to factor a dense probability table into a product of smaller tables. Even when the graph designer does not specify deterministic variables, methods have been developed to discover these factorizations [31]. In addition to their computational advantages, they are often a powerful representational tool in that deterministic variables can have real meaning and their posterior probabilities $p(d|\text{evidence})$ are often needed for further "semantic" processing. Their usefulness has prompted a body of research on efficient calculations in mixed deterministic/stochastic graphs [10, 13, 11, 7, 20, 3].

Computing exact probabilistic quantities can be done using a junction tree [22, 27], as a search procedure [3, 10], or using a hybrid scheme [10]. All exact inference methods in one way or another explicitly or at least implicitly define at least one graph triangulation [17], and many modern techniques can take advantage of a "good" triangulation. An important question, however, is how should the quality of a given triangulation be quantitatively judged?

A common measure of triangulation quality is the size of the clique potentials formed in traditional junction tree message passing, usually called the state space or weight [22, 19, 32]. Specifically, the **state space** or **cardinality** of a vertex $v$, notated $|v|$, is the number of distinct values it may hold. The state space of a *clique*, $C$, holding vertices $v_1, v_2, ..., v_k$ is defined as $S(C) = \prod_{i=1}^{k} |v_i|$. Lastly, the state space of a graph, $G$, with maximal cliques $C_1, C_2, ..., C_k$ is defined as $S(G) = \sum_{i=1}^{k} S(C_i)$. This measure is not fully adequate to describe the utility of a triangulation in every situation. Varying methods of computing probabilities have different performances and a wide variety of time-space trade offs. Search methods can be run in constant memory and use triangulations (which are often value specific) only indirectly. In addition, combinations of deterministic functions and evidence form constraints that can create large numbers of zeros in the distribution. Methods such as zero compression [16] and constraint propagation [12, 3, 20] can be used to exploit this, but the costs of queries on such a distribution are difficult to determine without actually performing the computation. With this being said, state space can give a measure of the upper bound of the computational requirements. As will be seen, we obtain significant wall-clock speedups under this assumption using a modern probabilistic inference engine.

This leads to the question of what method should be used to exploit deterministic variables in inference. Here we consider *arbitrary* deterministic functions (other techniques exist for certain classes of functions, such as sums [23]). In this work we use a hybrid approach where a junction tree is formed, but a search is used to process the messages entering and leaving the individual cliques. This can take advantage of some of the constraints imposed by the joint effect of determinism and evidence, but a more basic optimization comes from the fact that the value of a deterministic variable can be uniquely determined given its parents.

For a given parent combination it contributes only a constant time factor to evaluate the function, and performing this calculation does not require zero compression or any form of constraint propagation. If a deterministic variable lives in a maximal clique that is missing one or more of its parents, its value can not be determined uniquely and it needs to be iterated just as if it were stochastic. We now modify our definition of clique state space to reflect this:

**Definition 1.** *The state space of a clique, $C$, with $v_1, ..., v_k$, and the set $\mathcal{D} = \{v | v$ is deterministic and parents of $v \in C\}$ is: $S(C) = \prod_{v \in C \setminus \mathcal{D}} |v|$.*

Given this new optimization criterion, we show (Section 3) that on graphs with deterministic variables, triangulations that only minimize treewidth can use unboundedly more computational resources than triangulations with large cliques.

## 2 BACKGROUND

In this section, we define much of the notation, terminology, and known theorems required in later sections. A **chord** is an edge connecting two non-consecutive vertices in a cycle. A graph is **triangulated** if it contains no chordless cycles of length greater than three. A **triangulation** of a graph $G = (V, E)$ is a (possibly empty) set of edges $F$ such that $E \cap F = \emptyset$ and the graph $T(G) = (V, E \cup F)$ is triangulated. The edges in $F$ are called **fill-in edges**. The term triangulation will also be used to mean the graph $T(G)$. Given a graph $G = (V, E)$, the **neighbors** of $v \in V$ are defined as $NE_G(v) = (w \in V | \{v, w\} \in E)$. A triangulation $F$ of graph $G = (V, E)$ is **minimal** if $G' = (V, E \cup F_0)$ is not triangulated for any $F_0 \subset F$. A edge $e$ is a **non-minimal edge** in $T_1(G) = (V, E \cup F)$ if $T_2(G) = (V, E \cup F \setminus e)$ is also triangulated. A **clique** is a set of vertices for which every vertex in the set is connected to every other vertex in the set. A **maximal clique** is a clique that is not a subset of some larger clique. The **treewidth** of a triangulated graph [2] is the size of its largest maximal clique minus 1. The treewidth of an arbitrary graph is the smallest width of all triangulations. A vertex is **simplicial** in the graph $G$ if $NE_G(v)$ form a complete set. Triangulated graphs with more than one node have at least two simplicial vertices.

**Vertex elimination** [28, 8, 9] is an algorithm that can be used to triangulate graphs. An **elimination order** is a bijection $\alpha : \{1, 2, ..., |V|\} \leftrightarrow V$. We use $\alpha(v)$ to denote the integer position of node $v$ in the ordering $\alpha$, and $\alpha^{-1}(i)$ to denote the vertex indexed by the integer $i$ in ordering $\alpha$. The **deficiency** [28] of a vertex $v$ in $G$ is: $D_G(v) = (\{u, w\} | \{v, u\} \in E, \{v, w\} \in E, \{u, w\} \notin E\})$. Given $G = (V, E)$, the $v$-**elimination** graph $G_v$ is defined by adding the edges $D_G(v)$ and then deleting $v$ and its incident edges from $G$. Creating $G_v$ from $G$ is known as eliminating the vertex $v$. The **elimination graph**, denoted as $\xi_\alpha(G)$, is the original graph $G$ with the addition of any edges added at each step in the elimination process (see [28], $MTE(G; \alpha)$). A **perfect ordering** is an ordering for which elimination adds no edges.

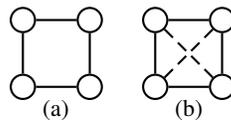

Figure 1: Triangulation (b) of (a) impossible with elimination [25]

All elimination graphs are triangulated [28], and a graph is triangulated if and only if it has a perfect elimination ordering [28, Theorem 1]. Some triangulations of a graph can not be generated via *any* elimination order ([25] and Figure 1). Elimination can, however, create any minimal triangulation [25]:

**Theorem 2.** *If $T(G) = (V, E \cup F)$ is a minimal triangulation of $G = (V, E)$, then there exists an elimination order $\alpha$ such that $\xi_\alpha(G) = T(G)$ [25, Theorem 1].*

The choice of triangulation can make an exponential difference in the time and memory needed for inference and finding an optimal one is NP-hard [2, 32], so heuristic search methods must be used. Many search methods exist, and are often based on choosing an ordering of the nodes for vertex elimination [18, 19, 24, 28]. Most heuristics that do not involve elimination (such as [6, 26]) will choose minimal triangulations over non-minimal triangulations.

## 3 MINIMAL VERSUS NON-MINIMAL

With a sense of the usefulness of deterministic variables and background on triangulation and elimination, we continue by discussing when and why one would want to move beyond conventional triangulation techniques. It is a common belief that triangulations that minimize clique size are desirable for use in computing queries on a graphical model. The reason is that it can be shown that many of the metrics of performance are upper bounded in some exponential function on $w$, the inherent treewidth of the graph [7, 13]. Although $w$ can form an upper bound it is *not* necessarily optimal. An instance of this is when it may be more desirable to cluster many small cardinality variables together in a clique to avoid increasing the sizes of cliques that contain large cardinality variables. In addition, some queries can be performed on junction trees without actually storing the clique potentials in memory; instead only the separator potentials are stored. In such a scheme one might want to increase clique size to reduce separator size and, in turn, reduce memory requirements (at the cost of more time) [11]. Also see [1] where large cliques are used to reduce junction tree clique degree. Later in this section it will be shown that large cliques can be beneficial when deterministic variables are present in the graph.

In graphs *without* deterministic variables, state space optimal triangulations might not always be treewidth optimal, but we show here that they will always be minimal and therefore obtainable using some elimination order. Many papers attempt to minimize state space by searching over elimination orders [19, 24] implying that this theorem has been assumed in the past, but we have not found a published proof in the literature. The proof of the following is given in the Appendix:

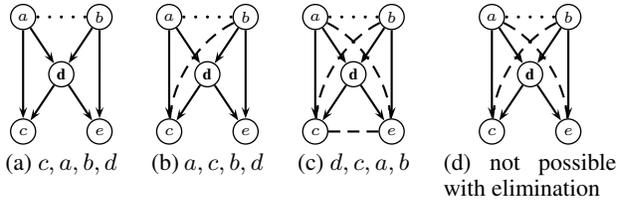

| (a) $c,a,b,d$ | (b) $a,c,b,d$ | (c) $d,c,a,b$ | (d) not possible with elimination |

Figure 2: Non-elimination based triangulation with unbounded improvement over elimination

**Theorem 3.** *Given a graph $G = (V,E)$ where all of the variables are stochastic and have state space $\geq 2$, some elimination graph of $G$ will have optimal state space.*

When using deterministic variables, state space optimal triangulations might not be obtainable from any elimination order. Consider Figure 2 where $d$ is a deterministic function of its parents $a$ and $b$, the cardinalities of $a,b,c$ and $e$ are all $\eta$, and the cardinality of $d$ is $\eta^2 - 1$ (the largest sensible cardinality for $d$). This graph is triangulated as is, and its state space with no additional fill-in is $2\eta^4 - \eta^2$. If one considers the graph in Figure 2(b) the cost is reduced to $\eta^4 + \eta^3 - \eta^2$. One might also run elimination beginning with $d$, resulting in the graph of Figure 2(c) and cost $\eta^4$. None of these nor any elimination ordering will give the optimal triangulation seen in Figure 2(d) having state space of $2\eta^3$. This state space is a factor of $\eta$ smaller than *any* elimination based triangulation. One might also notice that in this example the problem can be solved by transforming the graph into one that does not include the deterministic variable, where $a$ and $b$ are both connected directly $c$ and $e$. Standard elimination can then be used on the transformed graph. Although this approach could work, it will be shown in Section 5 and Figure 3 that the optimal choice of which transformations to make can not be made locally and is not any simpler than choosing a fill-in.

## 4 ELIMINATION GRAPH DETECTION AND COMPLEXITY

It has been demonstrated that elimination is unable to create certain (potentially useful) triangulations, but given a triangulation, how can one tell if an elimination order could have generated it? We give a polynomial time algorithm to solve this problem in Algorithm 1, and a correctness proof is given in the Appendix. It takes as input a graph, $G$, and a triangulation, $T(G)$, and returns true if $T(G)$ can be obtained by some elimination order. This algorithm is essential in our results where we show that most of the desirable triangulations in our test set could not have been generated by elimination.

Next, we give results on the computational complexity of finding triangulations in the mixed stochastic/deterministic setting. It was proven in [2] that it is NP-complete to determine whether a graph has treewidth $\leq k$. It was proven in [32] that finding optimal state space triangulations in graphical models with binary variables is NP-hard through a reduction from the Elimination Degree Sequence problem. Here we state that the decision version of the triangulation problem remains in NP for any polynomial time heuristic $f(T(G), I)$, where $I$ contains vertex information such as cardinality and determinism. This general definition allows us to prove NP-completeness for determining if a triangulation is sufficiently good under our modified definition of state space. We show that the state space problem remains NP-complete through a reduction from the treewidth problem. This reduction is simpler than the reduction from the Elimination Degree Sequence problem and is valid for variables with arbitrary cardinalities. See the Appendix for the proofs.

**Algorithm 1** isEliminationGraph

On input $\langle G = (V,E), T(G) = (V, E \cup F)\rangle$
**if** $|V| = 0$ **then**
  return *true*
**else**
  $A = \{v | (v \text{ simplicial in } T(G)) \ \& \ (\text{NE}_G(v) = \text{NE}_{T(G)}(v))\}$
  **if** $A = \emptyset$ **then**
    return *false*
  **else**
    $v \in A$, return isEliminationGraph( $G_v$, $(T(G))_v$ )
  **end if**
**end if**

**Definition 4.** MAXTRI $= \{\langle G = (V,E), I, \alpha\rangle \mid G$ has a triangulation with $f(T(G), I) < \alpha \}$

**Theorem 5.** *MAXTRI is in NP for all polynomial $f(G,I)$.*

**Definition 6.** MAXSTATESPACE $= \{\langle G = (V,E), I, \alpha\rangle \mid G$ has a $T(G)$ with state space $< \alpha\}$

**Theorem 7.** *MAXSTATESPACE is NP-complete.*

## 5 ELIMINATION WITH EXTRA FILL-IN

At this point, we know that elimination alone is not sufficient to generate all triangulations, and more importantly some graphs with deterministic dependencies can not be optimally triangulated by any elimination order. Because finding state space optimal triangulations is NP-hard, it is necessary to develop heuristic approaches that are able to find the desired non-minimal triangulations. This section describes an algorithm called extra-elimination that extends elimination to make it possible to find any triangulation. This algorithm is given here in its most general form, but at this point it has too much flexibility. Later in the paper we limit its options to make it a practical search algorithm.

**Definition 8. Extra-Elimination** *: Alternate the following two steps until no vertices remain: $(a)$ Add edges to the current graph, $(b)$ Eliminate a vertex. When finished, take the union of the extra edges added in the $(a)$ steps and the fill-in edges added in the $(b)$ steps and add them to the original graph.*

Extra-elimination allows extra edges to be added at any point during the elimination process (with the restriction that one is not allowed to add extra edges to a vertex that has already been eliminated), but the same triangulation will result regardless of when extra edges are added.

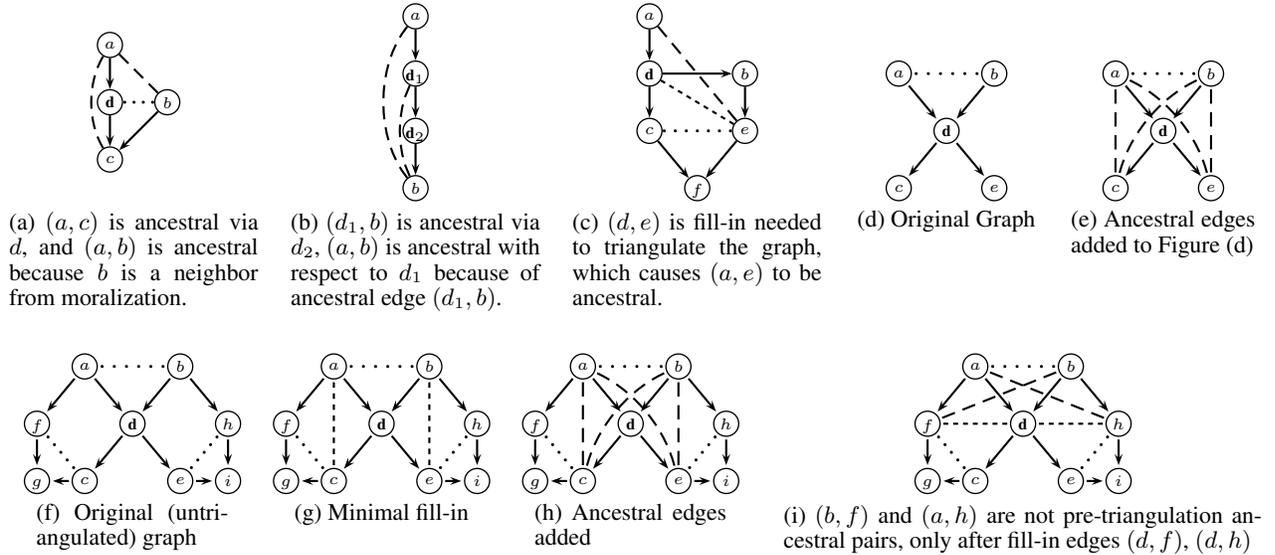

(a) $(a, c)$ is ancestral via $d$, and $(a, b)$ is ancestral because $b$ is a neighbor from moralization.

(b) $(d_1, b)$ is ancestral via $d_2$, $(a, b)$ is ancestral with respect to $d_1$ because of ancestral edge $(d_1, b)$.

(c) $(d, e)$ is fill-in needed to triangulate the graph, which causes $(a, e)$ to be ancestral.

(d) Original Graph

(e) Ancestral edges added to Figure (d)

(f) Original (untriangulated) graph

(g) Minimal fill-in

(h) Ancestral edges added

(i) $(b, f)$ and $(a, h)$ are not pre-triangulation ancestral pairs, only after fill-in edges $(d, f), (d, h)$

Figure 3: Figures (a)-(c) illustrate ancestral edges. Figures (d)-(i) illustrate why the choice of ancestral pairs can not be made locally. Variable **d** is deterministic in all graphs.

Extra-elimination can be considerably constrained by taking into account that we are considering Bayesian networks with deterministic variables that have values given by arbitrary functions of their parents. In such a graph, moralization will cause each deterministic variable to be in at least one maximal clique with its parents, but the variable might also be a member of other maximal cliques due to its non-parent neighbors. In these cases we can sometimes reduce the state space by adding fill-in edges that ensure that every maximal clique that contains a deterministic variable $d$ also contains $d$'s parents. We call edges that can accomplish this goal *ancestral edges* and these are the edges that we will pick from when choosing extra fill-in edges.

**Definition 9.** *An **ancestral edge** is one that connects a parent of a deterministic node $d$ to a child or undirected neighbor of $d$. An **ancestral pair** is a pair of nodes such that an edge between them would be ancestral.*

Ancestral edges are named for the case where they connect a node's parent to the node's child, but they can exist for a number of reasons. These include a deterministic node's undirected neighbors gained during moralization, undirected neighbors needed to triangulate the graph, or a neighbor from another ancestral edge (see Figures 3(a)-3(c)). In certain contexts there might be neighbors from other sources as well, such as DBN frame boundaries [4] or the creation of cliques for the analysis of posteriors or MAP calculations.

The following theorem states that the optimal state space triangulation can always be found using extra-elimination where we limit the choice of extra edges to ancestral pairs. The proof (in Appendix) shows that the state space optimal triangulation will be a minimal triangulation of a graph augmented by ancestral edges.

**Theorem 10.** *Elimination with extra-elimination edge addition where the extra edges are limited to ancestral edges is sufficient to find an optimal state space triangulation when all cardinalities are $\geq 2$.*

Our problem is still not solved, though. Theorem 10 only tells us that the needed non-minimal edges will be ancestral in the optimal triangulation. Not all ancestral edges may be needed, and not all needed ancestral edges will be known without knowing the rest of the optimal triangulation. It might at first seem that we can look at each deterministic node, $d$, with parents $\text{pa}(d)$, and non-parent neighbor $c$ and add ancestral edges between $\text{pa}(d)$ and $c$ if $S(c \cup d \cup \text{pa}(d))$ is less than $S(c \cup d) + S(d \cup \text{pa}(d))$. For example, in Figure 3(d) choosing ancestral pairs will only depend on the cardinality of $d$. Suppose $|a| = |b| = |c| = |e| = 10$ and $|d| = 40$, then the graph will have a state space of 900 with no fill-in, and a state space of 2000 when, as in Figure 3(e), all ancestral edges have been added. In the general case, the choice of ancestral pairs needed for a state space optimal triangulation can not be made without knowing the rest of the triangulation. In Figure 3(f), either $(a, c)$ or $(f, d)$ and either $(b, e)$ or $(d, h)$ needs to be added in order to triangulate the graph. Just as in Figure 3(d) it is not locally optimal to add the ancestral edges, but if $(a, c)$ and $(b, e)$ are added to triangulate the graph, as in Figure 3(g), the ancestral edges might now be beneficial. If the cardinality of all of the stochastic variables is 10 and the cardinality of $d$ is 40, then Figure 3(g) has a state space of 12100, but Figure 3(h) that includes the ancestral edges has a state space of only 6000.

Extra-elimination with ancestral edges gives a framework for finding triangulations in networks with deterministic dependencies, but gives no guidance on which ancestral pairs should be chosen. We now describe a pre-processing step that adds ancestral edges to a graph, and this new graph is then triangulated using standard elimination heuristics. The heuristics given here will only choose edges from what we will define as pre-triangulation ancestral pairs. A pair of vertices is a pre-triangulation ancestral pair if an edge between them is ancestral in the original moralized graph, or ancestral after other pre-triangulation ancestral edges have been added. Note that this definition implies a recursion: if an ancestral edge is added it might create additional ancestral pairs.

Four heuristics for deciding which extra edges to add are proposed here. The first is to use all pre-triangulation ancestral edges, called *all-extra*. The next is to randomly select a subset of pre-triangulation ancestral edges, called *sampled-extra*. The third heuristic is to choose the pre-triangulation extra edges that are locally optimal, called *locally-optimal-extra (lo-extra)*. That is, if we have a deterministic node, $v$, with parents $pa(v)$ and non-parent neighbor $c$, the set of edges between $c$ and $pa(v)$ is locally optimal if $S(c \cup v \cup pa(v)) < S(c \cup v) + S(v \cup pa(v))$. The fourth method is similar to all-extra, except that it ignores ancestral pairs that are a result of undirected edges (such as from moralization). It only considers ancestral pairs resulting from children of deterministic nodes plus any ancestral pairs that are recursively formed from other ancestral edge additions. This method is called *some-extra* and is included primarily to show the effect of ancestral edges due to these less obvious causes.

It should be stressed that considering only pre-triangulation ancestral edges does not consider all possible ancestral edge triangulations. The fill-in due to an elimination step could increase the potential number of ancestral edges. This is illustrated in Figure 3(i). To consider all possible ancestral pairs one would also need to recursively consider the additional ancestral pairs caused by the fill-in edges added after each elimination step.

## 5.1 RESULTS

The goal of the experiments is to compare the four extra-elimination heuristics to elimination. For each graph and each of the four extra-elimination heuristics, 19642 triangulations were generated and 488 of these were timed. The same procedure was repeated four times using pure elimination. There are two reasons for this repetition. First, it provides a fair comparison between elimination and the overall best of the four extra-elimination methods. Second, it ensures that a large part of the elimination search space has been explored. Note that the elimination triangulations have a significant advantage over the four extra-elimination methods individually since $4\times$ as many cases were considered.

The 19642 triangulations were generated using a variety of state of the art elimination heuristics. 20 one-step look ahead heuristics were used, including minimum weight, fill, size, and various combinations and repetitions of these. For each look ahead heuristic there was an additional parameter from 1-3 where the next node in the order is chosen randomly from the top $x$ choices. This parameter is similar to the Stochastic-Greedy Algorithm given in [15]. Maxi-mum cardinality search was also used, bringing the total to 61. For each of the 61 methods one triangulation was chosen for timing by having the lowest state space from separate pools of 100, 50, 10, and 1 triangulations. All of the above were repeated using a modified state space heuristic.

| | best | $<\times 2$ | $\times 2$–$\times 4$ | $\times 4$–$\times 8$ | $\times 8$–$\times 16$ | $\geq \times 16$ |
|---|---|---|---|---|---|---|
| **all-extra** | 240 | 98 | 14 | 1 | 3 | 0 |
| **sampled-extra** | 76 | 186 | 59 | 27 | 5 | 3 |
| **some-extra** | 50 | 204 | 71 | 21 | 7 | 3 |
| **lo-extra** | 18 | 63 | 58 | 68 | 54 | 95 |
| **elimination** | 15 | 43 | 58 | 66 | 56 | 118 |

Table 1: 'best' gives number of randomly generated graphs the method was the fastest. '$<\times 2$' gives number of graphs the method was not the best but took < twice the time of the best, etc.

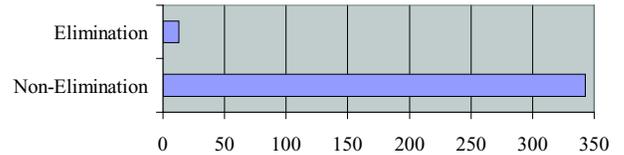

Figure 4: Number of random graphs where the fastest triangulation is an elimination graph versus not obtainable by elimination

The triangulations were timed on a calculation of the probability of evidence, stopping if more than 1 Gb of memory was used. Our probability computation engine is hybrid inference/search [10] where message passing is done over the junction tree and search is done within a clique. For search, we use an algorithm that consists of backtracking and an optimized static variable order.

The first set of graphs is composed of 356 randomly generated Bayesian networks. Each graph has 30 nodes, a maximum in-degree of 4, and the set of edges is chosen uniformly over all graphs fulfilling the constraints. Each node has a 0.5 probability of being deterministic and a 0.1 probability of being observed. The stochastic variables have cardinalities between 2 and 5 and the observed variables have a cardinality of 50. The deterministic variables have cardinalities between 2 and the product of their parents' cardinalities, with a upper bound of 125. For each graph, the fastest triangulation from each of the five methods is chosen. Table 1 gives counts of the number of graphs where each method was the best overall, and the number of times each method was various orders of magnitude slower than the best. Figure 4 compares the number of graphs where the best triangulation could and could not have been created using elimination (determined using Algorithm 1). Table 2 gives results over sets of graphs with a fixed number of deterministic variables. With little determinism, there is less opportunity for improvement over elimination, and with much determinism the total state space of the graph is small and the solution can be found quickly regardless of the triangulation.

The second set of results (Table 3) uses 10 real-world dynamic Bayesian networks. In addition to what was done on the random graphs, a set of 488 triangulations was generated using one instance of maximum cardinality search, minimum weight, fill, or size (labeled **once**). This is to compare our elimination baseline to a typical baseline. The following are descriptions of the graphs. **Aurora Decoding**: whole word model for speech recognition, [5]. **Edit Distance training 1, 2, decoding**: learns edit distance parameters from data [14]. **Feature Detect**: extracts phonetic features from speech data. **Image Concept Detect**: for image classification. **Mandarin**: speech recognition graph modeling tonal phones using asynchronous feature streams [33]. **MultiStream**: speech recognition training graph with asynchronous feature streams based on [34]. **PhoneFree 1, 2**: word pronunciation scoring using a phone-free model.

| # deterministic | 5 | 10 | 15 | 20 | 25 |
|---:|---:|---:|---:|---:|---:|
| all-extra | 82.8% | 93.5% | 100.0% | 87.5% | 100.0% |
| sampled-extra | 82.8 | 87.1 | 68.3 | 75.0 | 100.0 |
| some-extra | 62.1 | 71.0 | 70.7 | 75.0 | 100.0 |
| lo-extra | 44.8 | 51.6 | 17.1 | 50.0 | 89.3 |
| elimination | 58.6 | 35.5 | 9.8 | 54.2 | 71.4 |

Table 2: % of random graphs with fixed # of deterministic variables (out of 30) where method gave the shortest inference time or was $<2\times$ shortest.

On the randomly generated graphs, all-extra was the overall winner scoring the best on over half of the graphs. Sampled-extra was the second best overall, followed by some-extra. Lo-extra and elimination performed poorly overall. All-extra performs well when there is a high percentage of determinism (as in this set of random graphs). One might conclude that all-extra is the only method that should ever be considered, but in one case it was 15 times as slow as the best (which was an elimination graph). Sampled-extra has the potential to perform very well as it subsumes all of the other methods, but the large number of fill-in choices keep it, on average, slower than all-extra. The results on the real-world DBNs were much less dramatic. This because the cliques that are necessarily formed when partitioning the DBNs can account for a majority of the compute time and make the graphs fairly dense to begin with (see [4]). The extra-elimination heuristics gave significant improvement on 4 graphs with 2 more than doubling in speed. The median performance of the new heuristics was much better in many cases, but poor in others.

## 6 CONCLUSION

This paper has shown that large clique triangulations can be computationally useful on graphs containing deterministic variables. An example was given where the optimal triangulation has a state space that is arbitrarily smaller than all elimination based triangulations. An algorithm was presented to determine if a triangulation could have been generated using elimination, and it was shown that the generalized triangulation problem is NP-complete. Extra-elimination was introduced as a framework for producing any triangulation, and it was proven that extra edges can be limited to the ancestral edges when optimizing for state space. Novel heuristics based on ancestral edges were presented and results were given on randomly generated and real world graphs. Future work will include a joint search for triangulation and within-clique dynamic variable orderings for use in hybrid inference/search procedures.

## A APPENDIX

**Lemma 11.** *If $G = (V, E), G' = (V, E')$ triangulated with $E' \subseteq E, |E \setminus E'| = k$ then there is an increasing sequence $G' = G_0 \subset ... \subset G_k = G$ of triangulated graphs that differ by exactly one edge. [21, Lemma 2.21, page 20]*

**Lemma 12.** *Consider $T(G) = (V, E \cup F)$ with non-minimal edge $(u,v)$ and all $v \in C$ are stochastic. $C$ is the (necessarily) unique maximal clique in $T(G)$ that contains $u$ and $v$ in $T'(G) = (V, E \cup F \setminus (u,v))$, $C$ is split into cliques $C_u = C \setminus \{u\}$ and $C_v = C \setminus \{v\}$. Define $c = S(C \setminus \{u,v\})$, if $C_u$ and $C_v$ are maximal cliques in $T'(G)$ then $S(T'(G)) - S(T(G))$ will be $c(|u| + |v| - |u||v|)$. If $C_u$ is a subset of a maximal clique it is $(1 - |u|)c|v|$, if $C_v$ is a subset it is $(1 - |v|)c|u|$, and if both are it is $-c|u||v|$.*
*Proof.* The cliques not containing both $u$ and $v$ will be unaffected by the edge removal. $S(C) = c|u||v|$, and $S(C_v) = c|v|$ and $S(C_u) = c|u|$. □

**Lemma 13.** *In a graph where all variables are stochastic with cardinality $\geq 2$, a triangulation that is state space optimal and minimal will exist.*
*Proof.* Suppose we have a non-minimal triangulation $T(G)$ and remove an edge $(u,v)$ creating $T'(G)$, from Lemma 13 $S(T'(G)) - S(T(G)) < 0$. From Lemma 11, we can create a sequence from any $T(G)$ to any minimal $T'(G)$, and the state space of each graph will be $<=$ the previous graph. □

*Proof of Theorem 3.* From Lemma 13 and Theorem 2. □

The following lemma is assumed by [25] but never proven:

**Lemma 14.** *If $G_1 = (V, E)$, $G_2 = (V, E \cup E_2)$, $\xi_\alpha(G_1) = (V, E \cup F_1)$, and $\xi_\alpha(G_2) = (V, E \cup E_2 \cup F_2)$, then $F_1 \subseteq (E_2 \cup F_2)$.*
*Proof.* Obvious for $|V| = 1$, assume it is true for $|V| = N - 1$ and consider $|V| = N$. $\text{NE}_{G_1}(\alpha^{-1}(1)) \subseteq \text{NE}_{G_2}(\alpha^{-1}(1))$, so $D_{G_1}(\alpha^{-1}(1)) \subseteq (D_{G_2}(\alpha^{-1}(1)) \cup E_2)$. Therefore $(G_1)_{\alpha^{-1}(1)}$ is a spanning subgraph of $(G_2)_{\alpha^{-1}(1)}$, and a proof by the induction hypothesis. □

**Theorem 15.** *If $G = (V, E)$ with ordering $\alpha$, then $\{v, w\}$ is a fill-in of $\xi_\alpha$ iff $\exists$ a chain $[v, v_1, v_2, ..., v_k, w]$ in $G$ such that $\alpha(v_i) < \min(\alpha(v), \alpha(w)) \forall i = 1...k$ [29, Lemma 4].*

**Lemma 16.** *$\mathcal{E}_\alpha(G) = T(G)$, $\alpha(k)$ is simplicial in $T(G)$, and $\text{NE}_G(\alpha(k)) = \text{NE}_{T(G)}(\alpha(k))$, then $\beta = (\alpha(k), \alpha(1), \alpha(2), ..., \alpha(k-1), \alpha(k+1), ..., \alpha(|V|))$ and $\mathcal{E}_\beta(G) = T(G)$.*
*Proof.* Eliminate $T(G)$ according to $\beta$. $\alpha(k)$ is simplicial in $T(G)$ so eliminating it will not add edges. For $i = 1, ..., k-1, k+1, ..., |V|$, $\alpha(i)$ will have the same neighbors eliminating according to $\beta$ as it does eliminating according to $\alpha$. Therefore, $\beta$ is a perfect ordering of $T(G)$. $\mathcal{E}_\beta(G) = (V, E \cup F_\beta)$, $\mathcal{E}_\beta(T(G)) = (V, E \cup F)$ from Lemma 14, $F_\beta \subseteq F$.

Now suppose edge $(v, w) \in F$ but $(v, w) \notin F_\beta$. $\text{NE}_G(\alpha(k)) \subseteq \text{NE}_{\mathcal{E}_\beta(G)}(\alpha(k)) \subseteq \text{NE}_{T(G)}(\alpha(k)) = \text{NE}_G(\alpha(k))$, so $v, w \neq \alpha(k)$. Define $S_\alpha = \{u \in V :$

|  |  | Aurora Decode | Edit Dist. Train 1 | Edit Dist. Train 2 | Edit Dist. Decode | Feature Detect | Image Detect | Mandarin | Multi-Stream | Phone Free 1 | Phone Free 2 |
|---|---|---|---|---|---|---|---|---|---|---|---|
| **Once** | % Fail | %0.0 | %2.0 | %20.3 | %17.8 | %0.0 | %0.0 | %42.6 | %0.0 | %0.0 | %25.4 |
|  | $\mu \pm \sigma$ | 0.9±0.1 | 36.5±36.1 | 39.2±13.1 | 1240.0±2714.8 | 618.5±293.1 | 44.0±1.6 | 12.9±0.2 | 6.4±2.6 | 4.9±0.4 | 34.6±10.9 |
|  | Median | 0.9 | 36.0 | 36.1 | 96.5 | 576.4 | 43.5 | 13.1 | 5.6 | 5.0 | 45.1 |
|  | Best | 0.8 | 11.0 | 30.6 | 24.4 | 378.8 | 41.9 | 12.6 | 5.2 | 3.5 | 9.8 |
| **Elimination** | % Fail | %0.0 | %5.9 | %8.3 | %4.7 | %22.3 | %4.9 | %34.0 | %3.8 | %15.7 | %18.4 |
|  | $\mu \pm \sigma$ | 1.3±5.1 | 18.5±79.7 | 83.4±294.6 | 264.4±1144.4 | 495.9±222.7 | 48.6±49.9 | 12.8±10.1 | 20.4±118.7 | 138.5±1062.8 | 106.8±1081.5 |
|  | Median | 0.8 | 11.7 | 40.5 | 28.8 | 527.7 | 42.4 | 12.9 | 5.5 | 4.8 | 29.6 |
|  | Best | **0.2** | **2.8** | **4.6** | **1.0** | 12.2 | **28.5** | **6.8** | **3.0** | 2.2 | 4.2 |
| **All** | % Fail | %0.0 | %8.8 | %85.0 | %88.7 | %39.8 | %74.8 | %3.3 | %1.2 | %80.3 | %54.1 |
|  | $\mu \pm \sigma$ | 0.4±1.3 | 11.4±35.9 | 57.5±366.1 | 221.8±784.7 | 646.9±216.7 | 48.5±86.9 | 8.4±5.5 | 9.1±59.0 | 768.7±2749.0 | 517.9±2020.7 |
|  | Median | 0.2 | 6.6 | — | — | 1011.5 | — | 7.6 | 3.9 | — | — |
|  | Best | **0.2** | **2.8** | 6.2 | **1.0** | 212.9 | 29.2 | 7.4 | 3.6 | 3.7 | **1.9** |
| **Samp.** | % Fail | %0.0 | %6.0 | %34.4 | %27.3 | %61.1 | %22.4 | %12.7 | %4.2 | %44.7 | %53.5 |
|  | $\mu \pm \sigma$ | 1.4±10.1 | 11.5±9.8 | 356.3±1065.6 | 387.7±1145.8 | 464.6±518.9 | 102.4±63.7 | 10.1±8.9 | 19.2±71.2 | 1822.1±2737.6 | 2145.6±3351.7 |
|  | Median | 0.2 | 10.6 | 256.7 | 193.4 | — | 123.7 | 9.6 | 5.7 | 4633.9 | — |
|  | Best | **0.2** | **2.8** | 4.8 | **1.0** | 18.2 | 29.1 | **6.8** | **3.0** | 6.5 | 33.4 |
| **Some** | % Fail | %0.0 | %3.9 | %66.0 | %66.2 | %8.8 | %2.9 | %4.1 | %2.6 | %35.5 | %36.5 |
|  | $\mu \pm \sigma$ | 0.3±0.9 | 19.9±133.1 | 28.9±52.2 | 117.6±462.1 | 108.3±192.0 | 327.8±228.8 | 8.3±3.6 | 34.2±75.4 | 165.8±1371.0 | 169.9±1607.4 |
|  | Median | 0.2 | 6.8 | — | — | 31.1 | 356.0 | 7.6 | 10.7 | 14.3 | 27.7 |
|  | Best | **0.2** | **2.8** | 6.2 | **1.0** | 12.7 | 29.2 | 7.4 | **3.0** | 2.5 | 3.5 |
| **L.O.** | % Fail | %0.0 | %4.1 | %55.7 | %55.5 | %30.5 | %9.2 | %2.5 | %0.8 | %38.1 | %38.1 |
|  | $\mu \pm \sigma$ | 0.4±1.4 | 9.1±3.9 | 298.1±1272.3 | 102.4±387.0 | 140.7±267.0 | 45.7±20.2 | 8.0±3.4 | 5.3±14.0 | 61.1±262.2 | 23.2±4.7 |
|  | Median | 0.2 | 7.3 | — | — | 130.1 | 42.9 | 7.0 | 3.6 | 39.2 | 23.9 |
|  | Best | **0.2** | **2.8** | **4.5** | **1.0** | 8.9 | **28.8** | **6.8** | **2.9** | **1.0** | 2.4 |
| **Median once / elim.** | | 1.04 | 3.09 | 0.89 | 3.35 | 1.09 | 1.03 | 1.01 | 1.02 | 1.04 | 1.53 |
| **Median once / all** | | 4.32 | 5.47 | — | — | 0.57 | — | 1.72 | 1.44 | — | — |
| **Median once / samp.** | | 3.94 | 3.40 | 0.14 | 0.50 | — | 0.35 | 1.36 | 0.98 | 0.00 | — |
| **Median once / some** | | 4.45 | 5.32 | — | — | 18.51 | 0.12 | 1.72 | 0.52 | 0.35 | 1.63 |
| **Median once / L.O.** | | 3.90 | 4.90 | — | — | 4.43 | 1.02 | 1.87 | 1.54 | 0.13 | 1.89 |
| **Best once / Best extra** | | 5.02 | 3.94 | 6.75 | 25.14 | 42.39 | 1.46 | 1.86 | 1.78 | 3.57 | 5.11 |
| **Median elim. / all** | | 4.17 | 1.77 | — | — | 0.52 | — | 1.70 | 1.41 | — | — |
| **Median elim. / samp.** | | 3.80 | 1.10 | 0.16 | 0.15 | — | 0.34 | 1.34 | 0.96 | 0.00 | — |
| **Median elim. / some** | | 4.30 | 1.72 | — | — | 16.94 | 0.12 | 1.70 | 0.51 | 0.34 | 1.07 |
| **Median elim. / L.O.** | | 3.77 | 1.59 | — | — | 4.06 | 0.99 | 1.84 | 1.51 | 0.12 | 1.24 |
| **Best elim. / Best extra** | | 1.16 | 1.00 | 1.01 | 1.00 | 1.37 | 0.99 | 1.00 | 1.02 | 2.29 | 2.21 |
| **Best Is Elim.** | | no | yes | yes | yes | no | yes | yes | yes | no | no |

Table 3: Amount of time in seconds to compute 100 DBN frames. Given are the % of timed triangulations that finished within time/memory limits, mean, standard deviation, median, and best of the timings that finished. Ratios of the median and best of once and elimination are also given. The last row tells if elimination could have produced the best overall triangulation (or best within error).

$\alpha(u) < \min[\alpha(v), \alpha(w)]\}$ and $S_\beta = \{u \in V : \beta(u) < \min[\beta(v), \beta(w)]\}$. If $k < \min[\alpha(v), \alpha(w)]$ then $S_\alpha = S_\beta$. If $k > \min[\alpha(v), \alpha(w)]$ then $S_\alpha = S_\beta \setminus \alpha(k)$. From Theorem 15, there is a path in $G$, $[v, v_1, v_2, ..., v_l, w]$, where $\alpha(v_i) \in S_\alpha \forall i = 1...l$ — but since $(v, w) \notin F_\beta$, no such path $[v, v_1, v_2, ..., v_m, w]$ exists in $G$ such that $\beta(v_i) \in S_\beta \forall i = 1...m$. This, however, is a contradiction since $S_\alpha \subseteq S_\beta$. Therefore, we must have that $(v, w) \in F_\beta$. Therefore, $F_\beta = F$. □

**Theorem 17.** *isEliminationGraph will return true if and only if F can be generated by some elimination order $\alpha$.*

*Proof.* First we will show if return= *true* then $\exists \alpha$ such that $\mathcal{E}_\alpha = T(G)$. The proof is by induction on $|V|$. It holds for $|V| = 1$, assume it is true when $|V| = N - 1$, and consider when $|V| = N$. We assumed return=*true* on $\langle G, T(G) \rangle$, so return=*true* on $\langle G_v, (T(G))_v \rangle$, and from the induction hypothesis $\exists \beta$ s.t. $\mathcal{E}_{\beta(G_v)} = (T(G))_v$. Construct $\alpha$ by concatenating $v$ to the front of $\beta$. $\text{NE}_G(v) = \text{NE}_{T(G)}(v)$ so $\text{NE}_{\mathcal{E}_\alpha(G)}(v) = \text{NE}_{T(G)}(v)$.

Next we show if $\xi_\alpha = (V, E \cup F)$ then isEliminationGraph will return true. This will be proven by induction on $|V|$. Obvious for $|V| = 1$, assume it is true when $|V| = N - 1$, and consider when $|V| = N$. We use Lemma 16 to construct $\beta$ from $\alpha$. $G_{\alpha(k)}$ is eliminated in the order $\beta \setminus \alpha(k)$ it will generate $T(G)_{\alpha(k)}$, and from the induction hypothesis isEliminationGraph will return true. □

*Proof of Theorem 5.* Verify a member of MAXTRI using $V =$"On input $\langle V, E, I, F, \alpha \rangle$: 1) Build $G = (V, E \cup F)$. 2) Check if $G$ is triangulated. 3) If $f(G, I) < \alpha$ *accept*; otherwise *reject*." $|F| < |V|^2$, testing if $G = (V, E \cup F)$ is triangulated is $\in P$ [29, 30]. □

*Proof of Theorem 7.* MAXTREEWIDTH = $\{\langle G = (V, E), k \rangle \mid G$ has treewidth $\leq k \}$, is NP-complete [2]. To reduce from MAXTREEWIDTH, give $v \in V$ cardinality $|V|$ and no determinism, and return MAXSTATESPACE($G, I, |V|^{k+1}$). There are $\leq |V| - 1$ maximal cliques in $T(G)$, so if $T(G)$ has a treewidth $\leq k$ we get a max. state space $(|V| - 1)|V|^k < |V|^{k+1}$, and MAXSTATESPACE will accept. If the treewidth of $G$ is $\geq k + 1$ the min. state space is $|V|^{k+1}$ and MAXSTATESPACE will reject. Similarly, if MAXSTATESPACE accepts the treewidth $\leq k$ and if it rejects the treewidth is $> k$. □

**Lemma 18.** *Consider $G, T(G)$ with cardinalities $\geq 2$. Suppose an edge $(p, c)$ is added to form $T_{new}(G)$. If $S(T_{new}(G)) < S(T(G))$ then $(p, c)$ is ancestral with respect to some deterministic node $d$.*

*Proof.* From lemma 12, if the state space decreases the maximal clique with $p, c$ in $T_{new}(G)$ can not have only stochastic variables. There must be deterministic variable $d$ and a maximal clique $C_1 \in T(G)$ s.t. $c, d \in C_1, p \notin C_1$ where $p$ is a parent of $d$, and the addition of $(p, c)$ creates a new maximal clique $C_2$ such that $C_1 \subset C_2$ and $p \in C_2$. From moralization $c$ can not be a parent of $n$, so it must be a child or undirected neighbor. □

*Proof of Theorem 10.* The optimal state space triangulation is $T(G) = (V, E \cup F)$. Define set $A$ as all edges $\in F$ and ancestral in $T(G)$, and $G' = (V, E \cup A)$. $T(G)$ is a triangulation of $G'$ with fill-in $F' = F \setminus A$. We want to conclude that $F'$ is a minimal triangulation of $G'$. Assume this is not true, and $\exists M \subset F'$ such that $T_{min}(G') = (V, (E \cup A) \cup M)$ is triangulated. From Lemma 11 there is an increasing sequence of graphs from $T_{min}(G')$ to $T(G)$. $S(T(G))$ is optimal so $S(T_{min}(G')) \geq S(T(G))$. If $S(T_{min}(G')) > S(T(G))$ at least one graph in the sequence must have a lower state space than the previous, but from Lemma 18 this is a contradiction. Therefore, $F'$ is a minimal triangulation of $G'$ and $T(G)$ is an elimination graph of $G'$. If $S(T_{min}(G')) = S(T(G))$, $T_{min}(G')$ is also an optimal triangulation and elimination graph. $\square$

## Acknowledgments

This work was supported by ONR MURI grant N000140510388 and by NSF grant IIS-0093430. *Feature Detect* graph courtesy of Simon King, *Image Concept* courtesy of Brock Pytlik, *PhoneFree 1 and 2* courtesy of Karen Livescu.